\documentclass[10pt, a4paper]{article}
\usepackage[utf8]{inputenc}
\usepackage{lrec}
\usepackage{graphicx}
\usepackage{tabularx}
\usepackage{booktabs}
\usepackage{makecell}
\usepackage{enumitem}
\usepackage{soul}

\usepackage{epstopdf}

\usepackage{hyperref}
\usepackage{xstring}

\newcolumntype{Y}{>{\centering\arraybackslash}X}
\setlist[enumerate]{noitemsep, topsep=0pt}

\title{Evaluation of Sentence Representations in Polish}

\name{Sławomir Dadas, Michał Perełkiewicz, Rafał Poświata}

\address{National Information Processing Institute, Warsaw, Poland \\
         \{sdadas, mperelkiewicz, rposwiata\}@opi.org.pl\\}

\abstract{
Methods for learning sentence representations have been actively developed in recent years. However, the lack of pre-trained models and datasets annotated at the sentence level has been a problem for low-resource languages such as Polish which led to less interest in applying these methods to language-specific tasks. In this study, we introduce two new Polish datasets for evaluating sentence embeddings and provide a comprehensive evaluation of eight sentence representation methods including Polish and multilingual models. We consider classic word embedding models, recently developed contextual embeddings and multilingual sentence encoders, showing strengths and weaknesses of specific approaches. We also examine different methods of aggregating word vectors into a single sentence vector. \\ 
\newline  
}

\begin{document}

\maketitleabstract

\section{Introduction}

In recent years, advances in representation learning for text have led to significant breakthroughs in the field of natural language processing (NLP). Word embeddings have been used with particular success in many linguistic tasks such as text classification, sequence tagging or machine translation. Despite the wide adoption of word representations, especially in combination with deep learning architectures, some tasks require encoding larger chunks of text: sentences, paragraphs or documents. Information retrieval and plagiarism detection are common examples of problems that deal with longer texts and demand efficient methods of comparing or searching in large collections of documents. For this reason, there is an interest in the development of sentence and document embeddings.

\subsection{Related work}

A straightforward and commonly used method for representing sentences is to aggregate the representations of words making up a sentence into a single vector. In this approach, existing word embedding methods can be used as a source of word vectors. Originally, the most popular choices for word embeddings were Word2Vec \cite{mikolov2013efficient,mikolov2013distributed}, GloVe \cite{pennington2014glove} or FastText \cite{bojanowski2017enriching}. More recently, it has been shown that neural language models trained on large corpora can be used to generate contextualized word representations that outperform static word embeddings in multiple linguistic tasks. Notable examples of such models include ELMo \cite{peters2018deep}, BERT \cite{devlin2019bert}, Flair \cite{akbik2018contextual}, ERNIE \cite{zhang2019ernie}, XLNet \cite{yang2019xlnet}. The sequence of word vectors is usually transformed into a single sentence vector using Bag-of-Words (BoW) approach by computing an arithmetic or weighted mean but more complex approaches have been explored as well. \newcite{arora2017a} proposed a method known as \emph{Smooth Inverse Frequency (SIF)} which consisted of weighting word vectors and removing the projection of first principal component from the resulting matrix of sentences. \newcite{shen2018baseline} performed an analysis of different aggregation techniques for word vectors and showed that concatenating mean and max pooled vector from the sequence of words might improve performance on some downstream tasks. 

Another group of methods uses a more direct approach to building sentence representations by employing sentence level optimization objectives. It usually involves training an encoder-decoder neural network where the encoder is a component responsible for generating the resulting sentence embeddings. Skip-Thought vectors \cite{kiros2015skipthought} is one of the first popular architectures of this type. It is a self-supervised model that given a sentence representation, tries to reconstruct the previous and the next sentence in a text document. This idea was later improved by other self-supervised models \cite{gan2017learning,logeswaran2018an}. Some sentence encoders such as Universal Sentence Encoder \cite{cer2018universal} or InferSent \cite{conneau2017supervised} are trained in a supervised way using Stanford Natural Language Inference (SNLI) - large manually annotated corpus of English sentence pairs. Other methods try to learn multilingual sentence representations by machine translation, using datasets of aligned sentence pairs in different languages \cite{schwenk2017learning,artetxe2018massively}. Availability of large multilingual corpora and recent improvements in automatic parallel corpus mining methods allowed to train models capable of encoding sentences in many, even low-resource, languages. LASER encoder introduced by \newcite{artetxe2018massively} can handle 93 languages, recently released multilingual version of Universal Sentence Encoder \cite{yang2019multilingual} supports 16 languages.

With the development of sentence representation methods, more datasets and tools for their evaluation started to appear. There are already several studies analyzing performance and linguistic properties of sentence representations in English \cite{adi2017finegrained,conneau2018senteval,perone2018evaluation,zhu2018exploring}. Recently published study \cite{krasnowska-kieras-wroblewska-2019-empirical} has aimed to investigate retention of linguistic information in sentence embeddings for English and Polish languages by means of probing methods and downstream tasks. This study shows that correlation between results for Polish and English are high for most of the downstream tasks and probing methods. However, for low-resource languages like Polish, shortage of datasets, tools and pre-trained models is a major obstacle to the analogous studies and there is still a need for a thorough research. Initiatives such as Cross-Lingual NLI Corpus (XNLI) \cite{conneau2018xnli} are welcomed but we still need more language-specific data in order to perform the evaluation of sentence embeddings for low-resource languages. In this study, we present two new sentence level datasets in Polish and evaluate a number of sentence embedding methods on five tasks including topic classification, sentiment analysis in two domains, textual entailment and relatedness. We hope that this work will reduce the gap between Polish and English in research concerning sentence representations.

\subsection{Contributions}
Our contributions are the following: \\
\begin{enumerate}
    \item We perform a systematic evaluation of several sentence representation methods on five downstream tasks in Polish. The evaluation includes methods based on the aggregation of word vectors, contextual embeddings from language models and multilingual sentence encoders.
    \item We share a manually translated version of SICK (Sentences Involving Compositional Knowledge) dataset, containing 10,000 pairs of sentences annotated with textual entailment labels and semantic relatedness scores.
    \item We share an automatically extracted dataset (8TAGS) of almost 50,000 sentences relating to 8 topics such as food, sport or medicine.
    \item We provide a tool based on SentEval evaluation toolkit by \newcite{conneau2018senteval}, adapted to support Polish models and datasets. We make the source code\footnote{https://github.com/sdadas/polish-sentence-evaluation} available, as well as all pre-trained models\footnote{https://github.com/sdadas/polish-nlp-resources} used in this study: Polish Word2Vec, GloVe, FastText and ELMo.
\end{enumerate}

\section{Methodology}
In this section, we describe datasets, tasks, and sentence embedding methods included in this study. First, we present three datasets, two of which have been prepared by us for this evaluation. Next, we provide a summary of sentence representation methods and our model pre-training procedure.

\subsection{Description of datasets and tasks}

\paragraph{Wroclaw Corpus of Consumer Reviews Sentiment (WCCRS)} A corpus created by Wroclaw University of Science and Technology consisting of text annotated with sentiment labels \cite{Kocon2019}. This dataset contains consumer reviews relating to four domains: hotels, medical services, products and education. Data for all domains was annotated at the document level and two domains - hotels and medical services - were additionally annotated at the sentence level. We use those two domains in our sentence evaluation experiments as a multiclass sentiment classification tasks: \emph{WCCRS Hotels} nad \emph{WCCRS Medicine}. In both cases, each sentence is assigned to one of four classes:  \emph{positive}, \emph{negative}, \emph{neutral} or \emph{ambiguous}. We report classification accuracy for both tasks.

\paragraph{Sentences Involving Compositional Knowledge (SICK)} This dataset is a manually translated version of popular English natural language inference (NLI) corpus consisting of 10,000 sentence pairs \cite{marelli2014a}. NLI is the task of determining whether one statement (premise) semantically entails other statement (hypothesis). Such relation can be classified as \emph{entailment} (if the first sentence entails second sentence), \emph{neutral} (the first statement does not determine the truth value of the second statement) or \emph{contradiction} (if the first sentence is true, the second is false). Additionally, the original SICK dataset contains semantic relatedness scores for the sentence pairs as real numbers ranging from 1 to 5. When translating the corpus to Polish, we tried to be as close as possible to the original meaning. In some cases, however, two different English sentences had an identical translation in Polish. Such instances were slightly modified in order to preserve both the meaning and the syntactic differences in sentence pair. For evaluation of sentence representations on SICK, we follow the usual approach used for English. We add textual entailment task (\emph{SICK-E}) as a multiclass classification problem, reporting the classification accuracy. For semantic relatedness (\emph{SICK-R}), we learn a model that tries to predict the probability distribution of relatedness scores, as in \newcite{conneau2018senteval}. For this task, a Pearson correlation between predicted and original scores is reported.

\paragraph{8TAGS} A corpus created specifically for this study as a more challenging example of sentence classification. It contains about 50,000 sentences annotated with 8 topic labels: \emph{film}, \emph{history}, \emph{food}, \emph{medicine}, \emph{motorization}, \emph{work}, \emph{sport} and \emph{technology}. This dataset was created automatically by extracting sentences from headlines and short descriptions of articles posted on Polish social networking site \emph{wykop.pl}. The service allows users to annotate articles with one or more tags (categories). Dataset represents a selection of article sentences from 8 popular categories. The resulting corpus contains cleaned and tokenized, unambiguous sentences (tagged with only one of the selected categories), and longer than 30 characters. For this task, classification accuracy is reported.

\subsection{Methods}
The evaluation was conducted on eight word and sentence representation methods. Models for four of them (Word2Vec, GloVe, FastText, ELMo) were pre-trained by us on a large corpus of Polish consisting of Polish Wikipedia and a collection of Polish books and articles, 1.5 billion tokens in total. For other methods, we used already pre-trained and publicly available models. In the case of static word embedding models (Word2Vec, GloVe, FastText) we used the same hyperparameters and training procedure. The dimensionality of vectors has been set to 300. The vocabulary contained only lemmatized words with 3 or more occurrences in the corpus plus all numbers from 0 to 10,000 and a set of pre-defined Polish forenames and last names. In the case of ELMo, we used the same hyperparameters as \newcite{peters2018deep} for the model with 4096 hidden units which was trained for 10 epochs.

\begin{table*}[t]
  \centering
  \begin{tabularx}{400pt}{lYYYYY}
    \toprule
    \makecell[l]{\textbf{Method}} & \makecell{\textbf{WCCRS}\\\textbf{Hotels}} & \makecell{\textbf{WCCRS}\\\textbf{Medicine}} & \makecell{\textbf{SICK-E}} & \makecell{\textbf{SICK-R}} & \makecell{\textbf{8TAGS}} \\
    \midrule
    \textbf{Word embeddings} & \multicolumn{5}{c}{} \\
    \midrule
    Random & 65.83 & 60.64 & 72.77 & 0.628 & 31.95 \\
    Word2Vec & 78.19 & \textbf{73.23} & \textbf{75.42} & 0.746 & \textbf{70.27} \\
    GloVe & 80.05 & 72.54 & 73.81 & \textbf{0.756} & 69.78 \\
    FastText & \textbf{80.31} & 72.64 & 75.19 & 0.729 & 69.24 \\
    \midrule
    \textbf{Language models} & \multicolumn{5}{c}{} \\
    \midrule
    ELMo (all) & \textbf{85.52} & \textbf{78.42} & 77.15 & \textbf{0.789} & \textbf{71.41} \\
    ELMo (top) & 83.20 & 78.17 & 74.05 & 0.756 & 71.41 \\
    Flair & 80.82 & 75.46 & \textbf{78.43} & 0.743 & 65.62 \\
    BERT & 76.83 & 72.54 & 73.83 & 0.698 & 65.05 \\ 
    \midrule
    \textbf{Sentence encoders} & \multicolumn{5}{c}{} \\
    \midrule
    LASER & \textbf{81.21} & \textbf{78.17} & \textbf{82.21} & 0.825 & 64.91 \\
    USE & 79.47 & 73.78 & 82.14 & \textbf{0.833} & \textbf{69.92} \\
    \bottomrule
  \end{tabularx}
  \caption{Evaluation of sentence representations on four classification tasks and one semantic relatedness task (\emph{SICK-R}). For classification, we report accuracy of each model. For semantic relatedness, Pearson correlation between true and predicted relatedness scores is reported. BERT, LASER and USE are multilingual models, other models are Polish only.}
  \label{tab:eval_all}
\end{table*}

For other models, we utilized the following pre-trained versions: 1) For Flair, \emph{pl-forward} and \emph{pl-backward} models available in the library were used. 2) For BERT, we used the official BERT Multilingual Cased model. 3) In the case of Universal Sentence Encoder, we used \emph{universal-sentence-encoder-multilingual-large} version from TensorFlow Hub. 4) For LASER, we used the official model from Facebook.

\subsubsection{Word embeddings}
In this section, we describe methods based on static word embeddings. The final sentence vector is computed as an arithmetic mean of all vectors in the sentence. It is important to note that training word embedding models such as Word2Vec does not produce word vectors normalized to unit length. This is usually not a problem for word based tasks but we noticed that it does affect the performance in our case where we expect each word in the sentence to contribute equally to the resulting sentence vector. Therefore, we normalize all vectors to unit length before computing the mean.

\paragraph{Random baseline} We compare the performance of other models to a simple baseline which is a word embedding model initialized with random vectors. In this method, each word in the vocabulary has randomly generated representation that does not encode any meaningful semantic information.

\paragraph{Word2Vec} A model described in \newcite{mikolov2013efficient} is one of the first successful applications of neural networks for generating distributed word representations. The model learns a semantic relationship between words in a fixed length context window. The paper proposed two different methods of training the model: skip-gram (predicting context words given target word) and continuous bag-of-words (predicting target word given its context). The original paper and a follow-up paper \cite{mikolov2013distributed} describe two efficient methods for computing training loss: negative sampling and hierarchical softmax. In this study, we evaluate the most commonly used version of Word2Vec - skip-gram with negative sampling.

\paragraph{GloVe} Global Vectors \cite{pennington2014glove} is a count based method for learning word embeddings. It works by constructing a matrix of weighted co-occurrences of words in a fixed context and then learns to estimate the co-occurrence scores from the dot product of word vectors. 

\paragraph{FastText} A model proposed by \newcite{bojanowski2017enriching} is an extension of Word2Vec \cite{mikolov2013efficient}. The idea of FastText is to jointly learn the representation of words and sub-word units defined as a set of fixed lengths character n-grams. In this model, the resulting embedding is defined as a sum of the original word vector and all n-gram vectors for this word. One interesting property of this model is the ability to estimate representations for out-of-vocabulary (OOV) words by combining their n-gram embeddings.

\subsubsection{Contextual word representations}
In this section, we describe methods for generating contextual word representations. This is an active area of research and all of the following approaches have been introduced only recently. Such approaches base on pre-training neural language model on a large text corpus. Hidden states of this neural model can be then used as word representations. These models often have many parameters which makes them computationally more expensive than using static word embeddings. For evaluation, we use the same approach to building sentence vector as in the case of static word embeddings: we compute a mean of all generated word representations in a given sentence.

\paragraph{ELMo} Embeddings from Language Models is a deep learning architecture introduced by \newcite{peters2018deep}. It is a language model with a character input utilizing three bidirectional LSTM layers. First layer is responsible for generating context independent representation of a word from its characters, next two layers are used for encoding word based contextual information. Authors suggest that using hidden states of all three layers as a word representation can be beneficial for some downstream tasks compared to using only the last layer. Following this advice, we evaluate two versions of ELMo embeddings: one using only the hidden state of the top layer and other using a concatenation of all three hidden states.

\paragraph{Flair} A simple character based language model proposed by \newcite{akbik2018contextual} which proved to be especially effective for sequence annotation tasks such as part-of-speech tagging or named entity recognition. Despite the fact that the model is fully character based, the authors use hidden states of the last LSTM layer at every white space position after each word as a representation for this word. To support bidirectionality, a separate backward language model is trained. For our experiments, we concatenate the outputs of forward and backward models to build the final word representation. 

\paragraph{BERT} Bidirectional Encoder Representations \cite{devlin2019bert} from Transformers is a model using the transformer architecture which was previously applied in sequence to sequence tasks. BERT, unlike traditional language models, is trained with two objectives: predicting missing words in a sentence and predicting whether one sentence comes after the other. 

\subsubsection{Sentence encoders}
Our evaluation includes two pre-trained multilingual sentence encoders supporting Polish language. In this section, we briefly describe those models.

\paragraph{LASER} Language-Agnostic SEntence Representations \cite{artetxe2018massively} is a multi-lingual sentence encoder architecture that learns sentence vectors by translating sentences from the source language to two target languages. The available pre-trained model supports 93 languages including Polish, and has been trained on a parallel corpus of 223 million sentences.

\paragraph{USE} Universal Sentence Encoder was first presented in \newcite{cer2018universal}. It uses a transformer architecture and the training procedure involves multi-task objectives that include both self-supervised and fully supervised tasks. First publicly available pre-trained models were English only but recently a multilingual version of USE supporting 16 languages including Polish \cite{yang2019multilingual} has been made available.

\section{Experiments}
In this section, we demonstrate the experimental results of sentence representation methods in Polish on downstream linguistic tasks. First, we discuss the results of unaltered versions of all the methods. Next, we demonstrate the effect of two aggregation techniques on word embedding models and discuss the performance impact of vector dimensionality.

\subsection{Evaluation of sentence representations}
Table \ref{tab:eval_all} shows accuracy scores for four multiclass classification tasks (\emph{WCCRS Hotels, WCCRS Medicine, SICK-E, 8TAGS}) and Pearson correlation for semantic relatedness task (\emph{SICK-R}). These results were obtained by training a neural network for 10 epochs with a sentence vector as an input, a single hidden layer of 50 units and softmax output layer. The exception was again \emph{SICK-R} task where the continuous relatedness score was transformed to a probability distribution of relatedness values and the network input was a concatenation of the difference of sentence vectors and their dot product. This approach was compatible with other evaluation studies for \emph{SICK-R} task in English.

Although there is no method with top performance in every task, there are three methods with strong results for more than half them. ELMo outperforms every other approach on three classification problems. This is an impressive result given that ELMo is not a typical sentence encoder and has been trained on word level objective. This allows us to believe that a sequential model based on ELMo (e.g. network with LSTM layer and attention or pooling mechanism) would achieve even better results for those tasks. On the other hand, we can see a clear performance difference between word level models and multilingual sentence encoders on SICK dataset, for natural language inference and semantic relatedness tasks. Both USE and LASER show a comparable performance, almost 5\% better than any word based model. For classification, we cannot decide which encoder is better. While LASER clearly outperforms USE in sentiment classification, it also clearly loses in topic classification. 

Performance of multilingual BERT model in Polish is below our expectations as it is worse than Polish static word embedding models. While multilingual model is more universal than language-specific model, it seems that the price for multilingualism is quite high in the case of BERT. Polish Flair model achieves reasonable results, being ranked just above word embedding models and below ELMo for most tasks. It even gets the best result in \emph{SICK-E} task among all language models. 

Word embedding models are close to each other without an evident winner. They can be used as a strong baseline and for the evaluated tasks their performance is not far from more computationally expensive language models. 

\subsection{Evaluation of word embedding aggregation techniques and dimensionality}
\begin{table*}[t]
  \centering
  \begin{tabularx}{400pt}{lYYYYY}
    \toprule
    \makecell[l]{\textbf{Method}} & \makecell{\textbf{WCCRS}\\\textbf{Hotels}} & \makecell{\textbf{WCCRS}\\\textbf{Medicine}} & \makecell{\textbf{SICK-E}} & \makecell{\textbf{SICK-R}} & \makecell{\textbf{8TAGS}} \\
    \midrule
    \textbf{Word2Vec} & \multicolumn{5}{c}{} \\
    \midrule
    Baseline & 78.19 & \textbf{73.23} & 75.42 & 0.746 & 70.27 \\
    SIF & 77.22 & 70.37 & 72.04 & 0.729 & \textbf{71.32} \\
    Max Pooling & \textbf{79.86} & 72.64 & \textbf{79.47} & \textbf{0.792} & 69.85 \\
    \midrule
    \textbf{GloVe} & \multicolumn{5}{c}{} \\
    \midrule
    Baseline & 80.05 & 72.54 & 73.81 & 0.756 & 69.78 \\
    SIF & \textbf{80.24} & 72.05 & 72.38 & 0.758 & \textbf{69.83} \\
    Max Pooling & 78.25 & \textbf{73.68} & \textbf{77.56} & \textbf{0.786} & 69.42 \\
    \midrule
    \textbf{FastText} & \multicolumn{5}{c}{} \\
    Baseline & \textbf{80.31} & \textbf{72.64} & 75.19 & 0.729 & 69.24 \\
    SIF & 78.51 & 70.47 & 65.39 & 0.721 & \textbf{70.08} \\
    Max Pooling & 80.18 & 72.10 & \textbf{78.50} & \textbf{0.785} & 68.41 \\
    \bottomrule
  \end{tabularx}
  \caption{Evaluation of aggregation techniques for word based sentence representations with fixed vector dimensionality (300). Baseline model uses simple averaging, \emph{SIF} is a method proposed by \protect\newcite{arora2017a}, \emph{Max Pooling} is a concatenation of arithmetic mean and max pooled vector from word embeddings.}
  \label{tab:eval_static}
\end{table*}

\begin{figure*}
  \centering
  \includegraphics[scale=0.46]{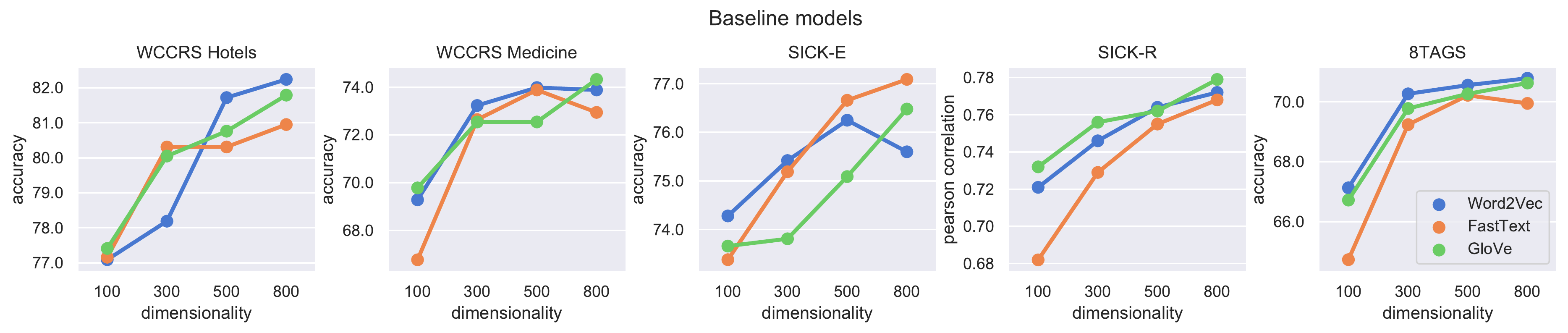}
  \includegraphics[scale=0.46]{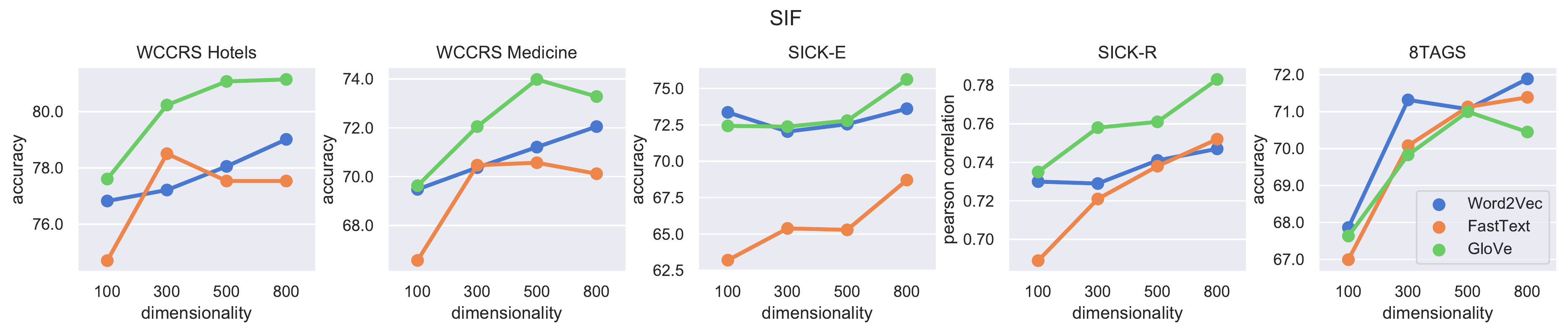}
  \includegraphics[scale=0.46]{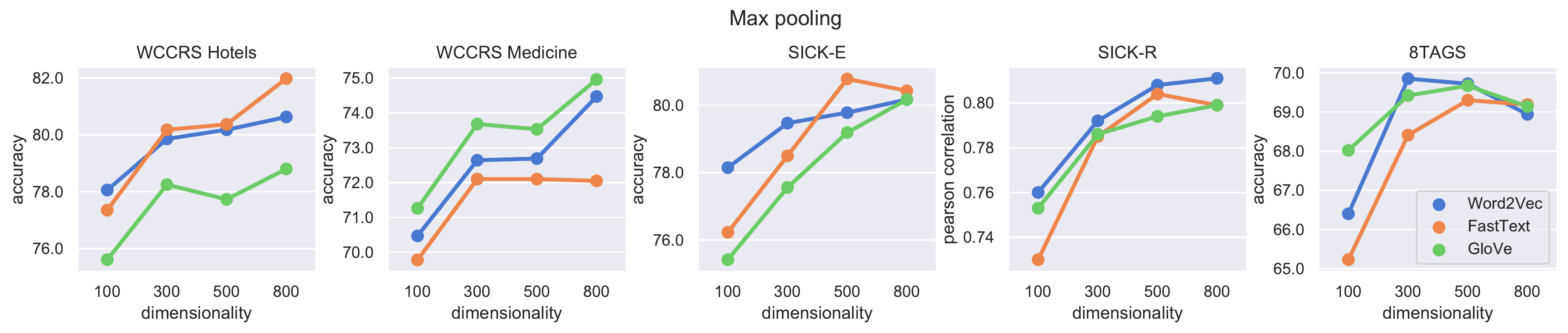}
  \caption{Evaluation of aggregation techniques for word embedding models with different dimensionalities. Baseline models use simple averaging, \emph{SIF} is a method proposed by \protect\newcite{arora2017a}, \emph{Max Pooling} is a concatenation of arithmetic mean and max pooled vector from word embeddings.}
  \label{fig:dim}
\end{figure*}

In the previous subsection, we compared the performance of various sentence representation methods, using an arithmetic mean to transform a sequence of word vectors into a single sentence vector. In this section, we examine two more advanced approaches to computing sentence representation from word vectors. First one, known as \emph{Smooth Inverse Frequency (SIF)} by \newcite{arora2017a}. It works by computing a weighted average of vectors with weights equal to $\frac{a}{a+p(w)}$, where $a$ is a smoothing parameter and $p(w)$ is a probability of the word, estimated as the frequency of the word in a corpus divided by the total number of tokens in this corpus. Then, a matrix of sentence embeddings is created and the first principal component of the resulting matrix is subtracted from it (\emph{common component removal}). The second evaluated method, described in \newcite{shen2018baseline}, involves concatenating simple average of word vectors and max pooled representation from the sequence of those vectors.

Table \ref{tab:eval_static} shows results of this experiment on classic word embedding models. Baseline is the same as in the previous experiment. We can see that the effect of applying \emph{SIF} method is not clearly beneficial. The performance is similar to the baseline model with small gains for some tasks and small losses for others. The authors of \emph{SIF} demonstrated that the method might improve semantic similarity of word vectors but it seems that there are no improvements for other tasks, at least in Polish. \emph{Max pooling}, however, turned out to beneficial for some problems. For each of the evaluated models, concatenating max pooled vector resulted in performance comparable to the baselines on classification tasks and significant improvements for tasks based on \emph{SICK} dataset, with gains of 3\% to 5\%.

We also studied the impact of the aforementioned aggregation techniques for word based models with different dimensionalities. In this experiment, we trained static word embeddings (Word2Vec, GloVe and FastText) with vector sizes of 100, 300, 500 and 800. Figure \ref{fig:dim} illustrates the performance of the baseline models as well as \emph{SIF} and \emph{max pooling} methods. We can see that simply increasing the size of the vector can improve accuracy. The largest models in their baseline versions turned out to almost as effective as ELMo on \emph{SICK-E}, \emph{SICK-R} and \emph{8TAGS} tasks. They also outperformed USE and LASER on \emph{WCCRS Hotels} and \emph{8TAGS}. This outcome is understandable as the more dimensions a word vector has, the more semantic information can be preserved in the resulting sentence representation. It is surprising, however, how close can these simple models get to the more sophisticated language models or sentence encoders.

The effect of applying \emph{SIF} and \emph{max pooling} is consistent with the previous experiment. Again, we see no clear improvement with \emph{SIF}. Application of \emph{max pooling} technique has a positive impact on \emph{SICK-E} and \emph{SICK-R} tasks, although it is more evident for smaller vectors. While this effect is partially mitigated in higher dimensions, the performance of the best word based model with \emph{max pooling} is just 1\% below multilingual sentence encoders on \emph{SICK}. Therefore, we can conclude that using a sufficiently highly dimensional word embedding model with \emph{max pooling} provides a strong baseline for sentence representations which is hard to beat event by the recent state-of-the-art methods.

\section{Conclusion}
In this study, we provided an experimental evaluation of sentence embedding methods for Polish on five linguistic tasks. We presented a comparison of three groups of methods: static word embeddings, contextual embeddings from language models and sentence encoders. We also examined two recently introduced methods for aggregating word vectors into a sentence vector. In order to accelerate research in Polish natural language processing, we make the source code for our experiments, our datasets and pre-trained models public.

\section{Bibliographical References}
\label{main:ref}

\bibliographystyle{lrec}
\bibliography{references}


\end{document}